\documentclass[letterpaper, 10 pt, journal, twoside]{IEEEtran}
\ifCLASSINFOpdf
  \usepackage[pdftex]{graphicx}
\else
\fi
\usepackage{amsmath}
\usepackage{booktabs}     % for table formatting
\usepackage{xcolor}       % for text colors
\usepackage{hyperref}     % for hyperlinks
\usepackage{algorithmic}  % for algorithms
\usepackage{algorithm}    % for algorithms
\usepackage{amsfonts}     % for \mathbb
\usepackage{cite}         % to render [1]-[3] instead of [1][2][3]

\newcommand{\norm}[1]{\left\lVert#1\right\rVert}
\begin{document}
\title{Approximating Global Contact-Implicit MPC \\ via Sampling and Local Complementarity}

% \author{Michael~Shell,~\IEEEmembership{Member,~IEEE,}
%         John~Doe,~\IEEEmembership{Fellow,~OSA,}
%         and~Jane~Doe,~\IEEEmembership{Life~Fellow,~IEEE}% <-this % stops a space
% \thanks{M. Shell was with the Department
% of Electrical and Computer Engineering, Georgia Institute of Technology, Atlanta,
% GA, 30332 USA e-mail: (see http://www.michaelshell.org/contact.html).}% <-this % stops a space
% \thanks{J. Doe and J. Doe are with Anonymous University.}% <-this % stops a space
% \thanks{Manuscript received April 19, 2005; revised August 26, 2015.}}
\author{
    Sharanya Venkatesh*, Bibit Bianchini*\textsuperscript{\textdagger}, Alp Aydinoglu, William Yang, Michael Posa\\ {*}The first two authors contributed equally to this work. \textsuperscript{\textdagger}Corresponding author ({\tt\small bibit@seas.upenn.edu}).%
    \thanks{Manuscript received: May 13, 2025; Revised August 4, 2025; Accepted September 5, 2025.
    This paper was recommended for publication by Editor Júlia Borràs Sol upon evaluation of the Associate Editor and Reviewers' comments.
    This work was supported by a National Defense Science and Engineering Graduate (NDSEG) Fellowship, an NSF CAREER Award (FRR-2238480), and the RAI Institute.} %Use only for final RAL version
    \thanks{S. Venkatesh, B. Bianchini, and M. Posa are with the GRASP Laboratory at the University of Pennsylvania, Philadelphia, PA 19104.}%
            % {\tt\footnotesize bibit@seas.upenn.edu}}%
    \thanks{A. Aydinoglu is with Boston Dynamics, Waltham, MA 02451.}%
    \thanks{W. Yang is with Amazon Robotics, Seattle, WA 98109.}%
    \thanks{Digital Object Identifier (DOI): see top of this page.}
}

\markboth{IEEE Robotics and Automation Letters. Preprint Version. Accepted September, 2025}
{Venkatesh, Bianchini \MakeLowercase{\textit{et al.}}: Approximating Global CI-MPC} 




\maketitle

\begin{abstract}
To achieve general-purpose dexterous manipulation, robots must rapidly devise and execute contact-rich behaviors.  Existing model-based controllers cannot globally optimize in real time over the exponential number of possible contact sequences. Instead, progress in contact-implicit control leverages simpler models that, while still hybrid, make local approximations. Locality limits the controller to exploit only nearby interactions, requiring intervention to richly explore contacts more broadly.  Our approach leverages the strengths of local complementarity-based control combined with low-dimensional, but global, sampling of possible end effector locations. Our key insight is to consider a contact-free stage preceding a contact-rich stage at every control loop. Our algorithm, in parallel, samples end effector locations to which the contact-free stage can move the robot, then considers the cost predicted by contact-rich MPC local to each sampled location. The result is a globally-informed, contact-implicit controller capable of real-time dexterous manipulation. We demonstrate our controller on precise, non-prehensile manipulation of non-convex objects with a Franka arm.  Project webpage: \href{https://approximating-global-ci-mpc.github.io}{https://approximating-global-ci-mpc.github.io}
\end{abstract}

% \begin{IEEEkeywords}
% IEEE, IEEEtran, journal, \LaTeX, paper, template.
% \end{IEEEkeywords}

% \fi
\IEEEpeerreviewmaketitle

\section{Introduction}
% for IEEE journal papers produced under \LaTeX\ using
% IEEEtran.cls version 1.8b and later.
\IEEEPARstart{F}{or} multi-purpose robots to successfully deploy into the home and workplace, they will need to be capable of dexterous manipulation of complex objects.  Progress towards this goal has been made on many fronts, including imitation learning from human demonstrations \cite{chi2023diffusion} and offline model-based planning \cite{cheng2023enhancing, pang2023global}.  However, these approaches both require advanced knowledge of the task and can fail to generalize to even minor permutations, such as new goal configurations.

An alternative approach that has made recent strides is contact-implicit model predictive control (CI-MPC).  Contact-implicit methods optimize for state and/or input trajectories and a corresponding sequence of contact modes.
Facing hybrid, nonlinear dynamics, CI-MPC approaches compromise to reach real-time rates, such as by picking one of many sampled trajectories \cite{li2024drop, howell2022predictive}, or using simplified dynamics.  One such simplified model that captures the hybrid aspect of contact-rich dynamics is a linear complementarity system (LCS) \cite{heemels2000linear}, a piecewise-linear representation.  However like many simplified models, LCS dynamics are limited in accuracy to a local neighborhood of the true underlying system (Fig. \ref{fig:lcs_cartoons}).

\begin{figure}[t]
    \centering
    \includegraphics[width=0.78\linewidth,trim={0mm, 22mm, 134mm, 45mm},clip]{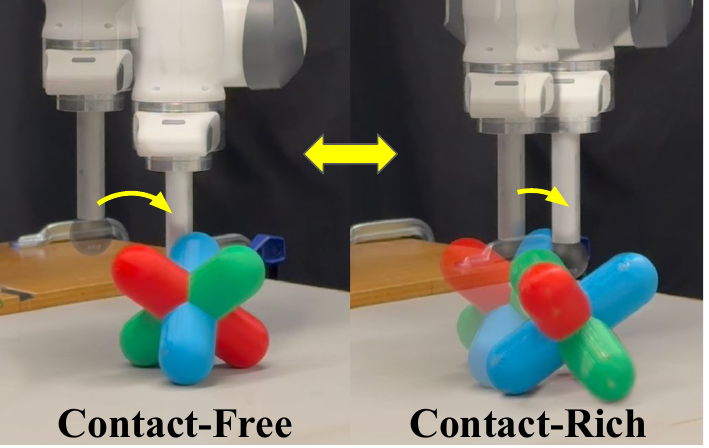}
    \vspace{-0.5em}
    \caption{Our real-time CI-MPC combines a global, explorative contact-free stage with a local contact-rich stage.  At every control loop, our algorithm chooses contact-rich actions to make goal progress, or contact-free actions to pursue more amenable starting locations for future contact-rich actions.}
    \label{fig:jack_experiment}
\end{figure}

Our novel CI-MPC approach inherits local efficacy of complementarity-based control and global efficacy of sampling-based strategies, only requiring sampling individual robot collision geometry locations.  This can be a tractably low-dimensional space, e.g. 3D end effector locations.  Our key insight is that all contact-rich control can be split into a contact-free stage when the robot moves in a collision-free path (easier to compute than contact-rich trajectories), followed by a contact-rich stage when the robot makes and breaks contact freely (Fig. \ref{fig:jack_experiment}).  Existing CI-MPC methods can be effective in the contact-rich stage, if the local neighborhood is amenable to making progress.  Sampling is a natural solution:  we sample end effector locations where the contact-free-to-rich transition can occur, quantifying sample advantages by their local CI-MPC optimization costs.  In closed loop with cost- and progress-based switching logic, our controller autonomously switches between contact-free and contact-rich modes, trading off future investment with immediate progress, respectively.

\begin{figure*}[t]
    \centering
    \includegraphics[width=\textwidth,trim={25mm, 57mm, 25mm, 55mm},clip]{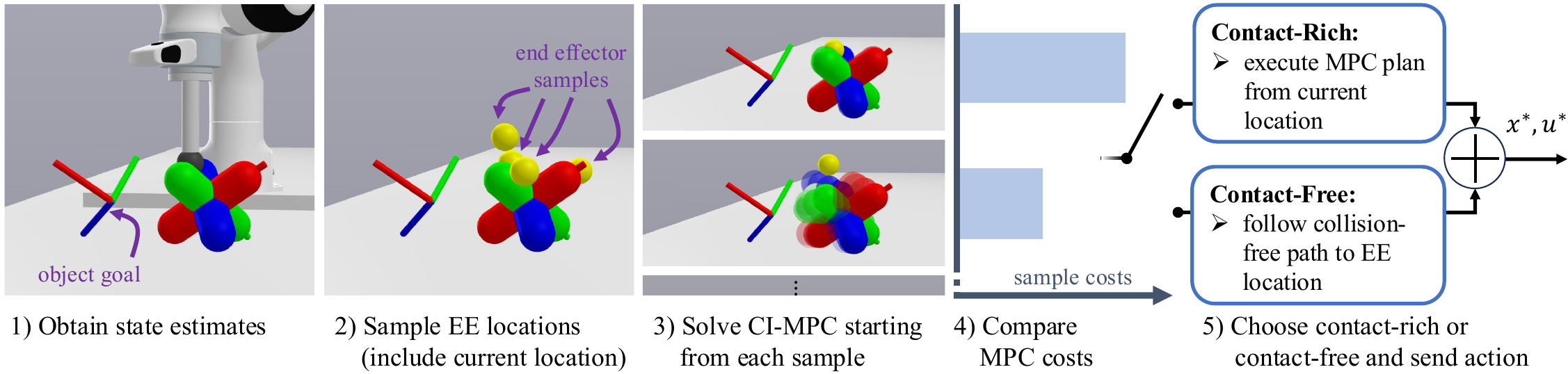}
    \vspace{-2em}
    \caption{The algorithm for one control loop of our sampling-based contact-implicit controller.  The third step that solves a local CI-MPC problem for each sample can be parallelized, since each plan is independent.  In this example, the top sample's CI-MPC plan makes little progress to the object goal, and thus is associated with high cost.  The second sample's CI-MPC plan shows more progress, and thus is lower cost.}
    \label{fig:algorithm}
\end{figure*}

In this paper, we contribute:
\begin{itemize}
    \item An online, multi-contact manipulation method combining global sampling with local control, requiring no offline computation or training.  Our approach accomplishes tasks that local control alone essentially always fails to do.
    \item Hardware results of our algorithm achieving pose goals of non-convex objects to high precision with a Franka arm, using 1 robot-object and 3 object-environment contacts.
    \item Ablations and comparisons in simulation, including to MuJoCo MPC (MJPC) \cite{howell2022predictive}, another non-local CI-MPC method our approach outperforms in success and safety.
\end{itemize}

%%%%%%%%%%%%%%%%%%%%%%%%%%%%%%%
\section{Related Work} \label{sec:lit}

\subsection{Offline Trajectory Optimization} \label{subsec:rel_control:offline}
Solving a global nonlinear control problem can be too time-intensive to compute online, but offline contact-implicit trajectory optimization can be useful \cite{contact_johnson_2019, posa2014direct}.  Non-smooth contact dynamics can create confront challenging cost landscapes \cite{antonova2023rethinking}.  Mitigations include artificially smoothing gradients \cite{pang2023global, suh2022bundled}, graph search \cite{cheng2023enhancing} or sampling \cite{zhu2023efficient} over contact mode sequences and/or control inputs \cite{liu2010sampling}, or restricting applications to simpler 2D problems \cite{aceituno2022hierarchical}.
Task and motion planning (TAMP) (e.g. \cite{toussaint2018differentiable}) is a flavor of making high-level choices combined with lower-level trajectory optimization.
There is a long history of combining global sampling and local optimization (in robotics, see \cite{antonova2023rethinking, kamat2022bitkomo}, among others).
Despite some applications to robotic manipulation \cite{chavan2019sampling}, to the best of our knowledge these methods have not achieved real-time contact-rich control.

\subsection{Contact-Implicit MPC} \label{subsec:rel_control:implicit}
Recent progress has reformulated global tasks into more tractable online MPC problems.  One approach uses local complementarity-based approximations to the global system \cite{yang2024dynamic, aydinoglu2024consensus}, encouraging consensus between local non-smooth rigid-body contact constraints and local dynamics, while exploring contact modes via small-scale mixed integer optimization. Other distinct but similar approaches use artificially smoothed contact dynamics, allowing online gradient-based algorithms to explore different contact modes \cite{kim2023contact, cleac2021fast, kurtz2023inverse, tassa2012synthesis}.
All of these methods, however, leverage local approximations and thus may fundamentally struggle to escape local minima and initiate distant, but beneficial, contact without intervention.

\subsection{The Role of Sampling for Planning and Control} \label{subsec:rel_control:sample}
While the non-smooth dynamics of contact-rich scenarios pose serious challenges to gradient-based techniques,
gradient-free methods like random sampling have shown promise.  Many methods in the literature include sampling in the spaces of input trajectories \cite{howell2022predictive, williams2016aggressive}, contact mode sequences \cite{xue2024full, zhu2023efficient}, or both \cite{liu2010sampling}.
In our work, we explore a lower-dimensional form of sampling:  sampling end effector configurations, using these as global seeds for the local contact-implicit controller from \cite{aydinoglu2024consensus}.  This does not require sampling full trajectories nor full system states, but retains the primary benefit of sampling as a way out of local minima.

%%%%%%%%%%%%%%%%%%%%%%%%%%%%%%%
\section{Background}
\label{sec:background}

We review modeling choices for contact dynamics (\S \ref{subsec:back:contact}), introduce a contact-rich optimal control problem (\S \ref{subsec:back:optimal}), then detail C3 \cite{aydinoglu2024consensus}, an existing real-time, contact-implicit
algorithm that locally approximates the control problem (\S \ref{subsec:back:c3}).

\subsection{Contact Dynamics}
\label{subsec:back:contact}

For a contact-rich system with state $x$ and inputs $u$ subject to contact forces $\lambda$, its dynamics can be generally written as
\begin{subequations} \label{eq:dynamics_comp}
    \begin{align}
        & x_{k+1} = g \left( x_k, u_k, \lambda_k \right), \label{eq:g} \\
        & 0 \leq \lambda_k \perp \pi \left( x_k, u_k, \lambda_k \right) \geq 0, \label{eq:pi}
    \end{align}
\end{subequations}
where the discrete-time dynamics $g$ depend on the contact forces $\lambda_k$, which are the solution to a nonlinear complementarity problem (NCP) \cite{stewart2011dynamics}
in \eqref{eq:pi}.  The NCP elegantly embeds the multi-modal nature of contact-rich systems.

In our context, the vector $\lambda_k$ represents contact forces, and the complementarity constraint \eqref{eq:pi}
enforces the hybrid (or non-smooth) aspects of the dynamics; for example, it can encode the constraint that forces must only occur when in contact, and can enforce the stick-slip effects of Coulomb friction.
For simulation, the solved contact forces are found to satisfy contact dynamics encoded by the NCP in \eqref{eq:pi} and affect the system dynamics as in \eqref{eq:g}.

A local approximation to \eqref{eq:dynamics_comp} that still preserves the multi-modality of contact is a linear complementarity system (LCS) \cite{heemels2000linear}.  
An LCS describes the state and contact force trajectories for an input sequence starting from $x_0$ such that
\begin{subequations} \label{eq:lcs}
\begin{align}
& x_{k+1} = A x_k + B u_k + D\lambda_k + d, \label{eq:lcs_dynamics} \\
& 0 \leq \lambda_k \perp Ex_k +  F \lambda_k + H u_k + c \geq 0. \label{eq:lcs_contact}
\end{align}
\end{subequations}
where $x_k \in \mathbb{R}^{n_x}$, $\lambda_k \in \mathbb{R}^{n_{\lambda}}$, $u_k \in \mathbb{R}^{n_u}$, $A \in \mathbb{R}^{n_x \times n_x}$, $B \in \mathbb{R}^{n_x \times n_u}$, $D \in \mathbb{R}^{n_x \times n_\lambda}$, $d \in \mathbb{R}^{n_x}$, $E \in \mathbb{R}^{n_{\lambda} \times n_x}$, $F \in \mathbb{R}^{n_\lambda \times n_\lambda}$, $H \in \mathbb{R}^{n_\lambda \times n_u}$, and $c \in \mathbb{R}^{n_\lambda}$.

In computing the LCS representation of a given system, we consider a nominal $x_\text{nom}, u_\text{nom}$ and the resulting $\lambda_\text{nom}$ from solving \eqref{eq:pi}.
The LCS dynamics \eqref{eq:lcs} approximate the non-linear hybrid dynamics \eqref{eq:dynamics_comp} by linearizing $g$ and $\pi$.

Once the LCS is constructed, for a given $x_k$ and $u_k$, the corresponding complementarity variable $\lambda_k$ is found by solving \eqref{eq:lcs_contact}. Similarly, $x_{k+1}$ can be computed using \eqref{eq:lcs_dynamics} when $x_k, u_k$, and $\lambda_k$ are known. Given an input $u_k$ and state $x_k$, the next state is $x_{k+1} = \text{LCS}(x_k,u_k)$ defined by \eqref{eq:lcs}.

\begin{figure*}[t]
    \centering
    \includegraphics[width=0.52\textwidth,trim={0mm, 30mm, 8mm, 30mm},clip]{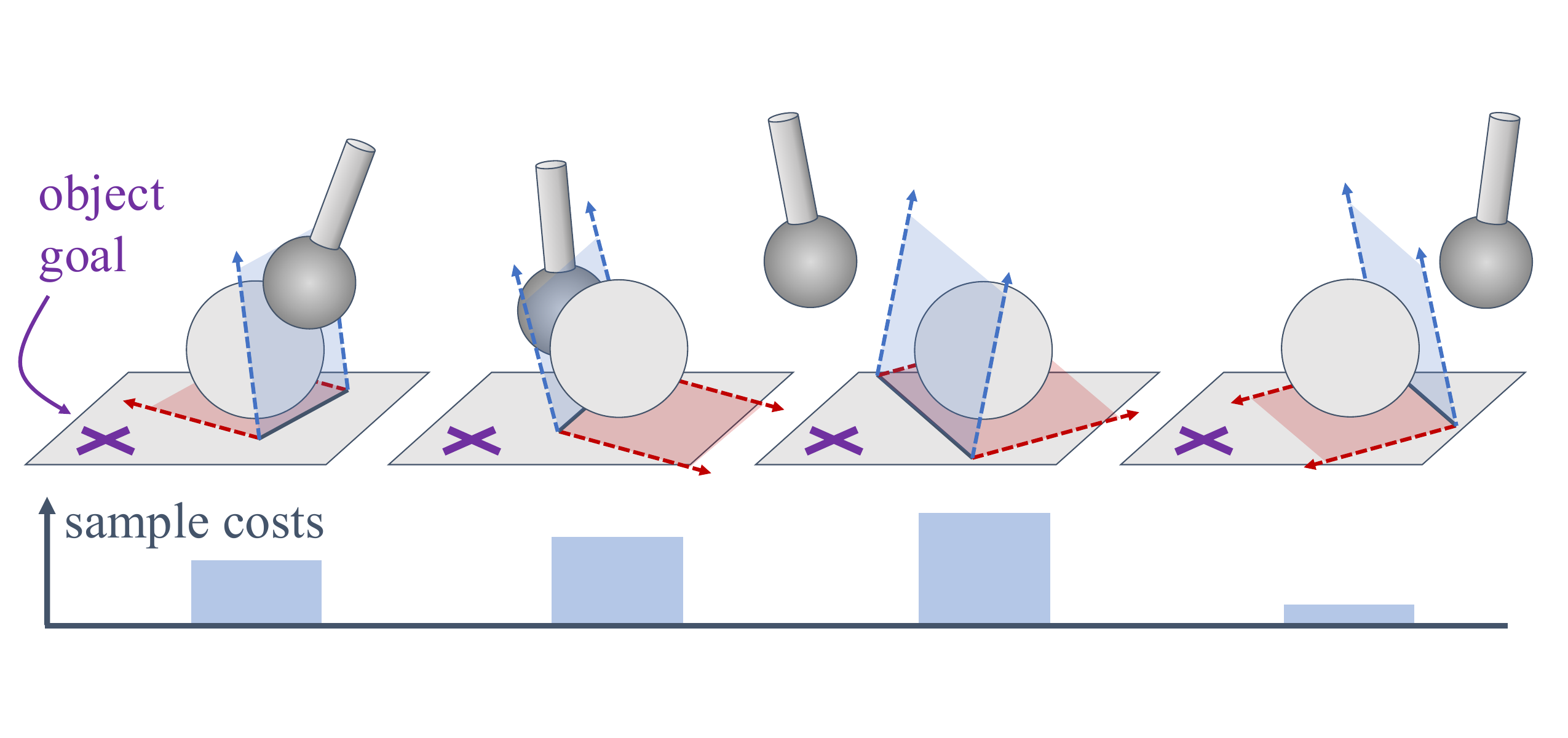}
    \quad
    \includegraphics[width=0.44\textwidth,trim={65mm, 32mm, 62mm, 32mm},clip]{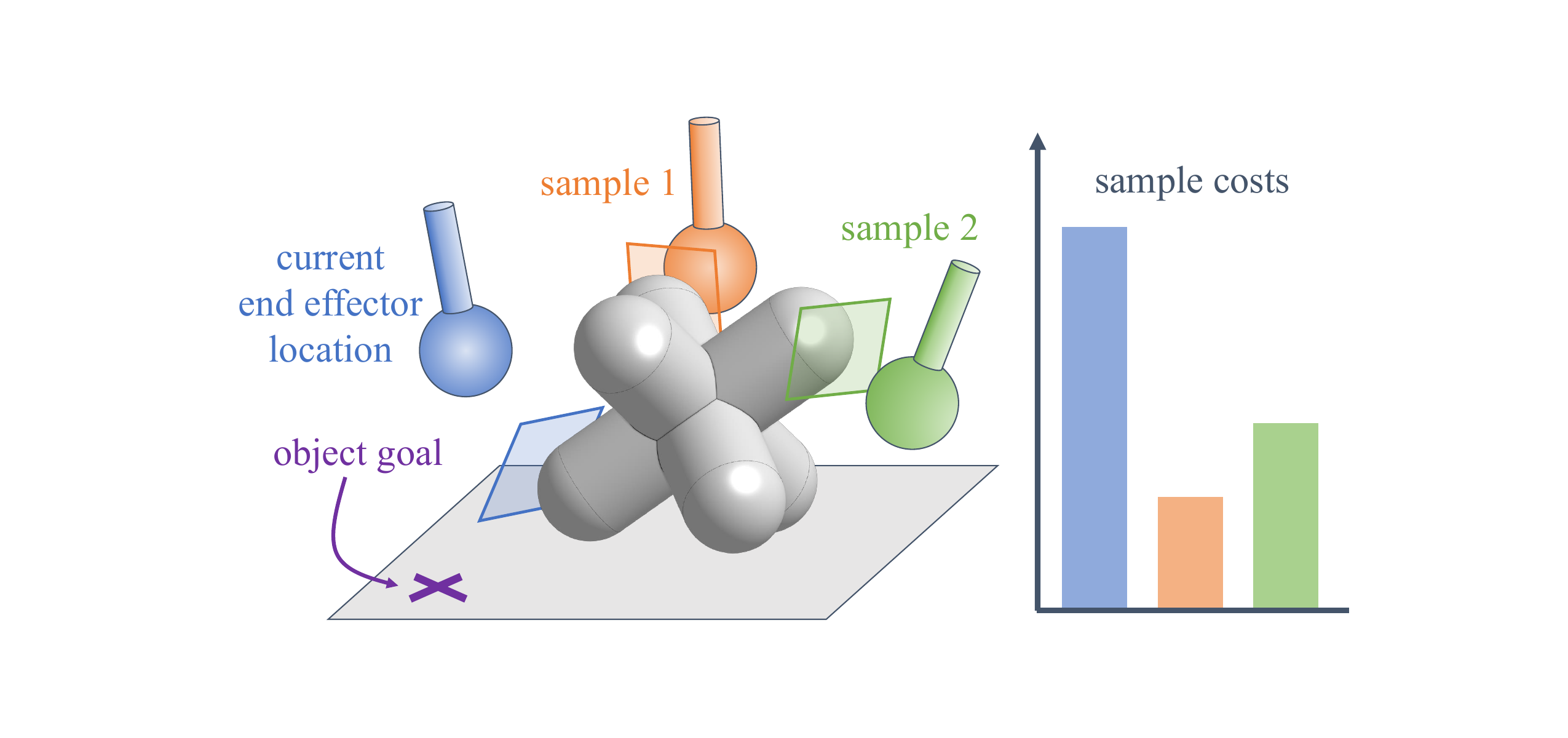}
    \caption{Left:  A spherical end effector approaches a spherical object on a flat table.  Loosely speaking, the LCS approximates object geometry as a set of hyperplanes coincident with and tangent to their witness points with respect to other geometries of interest.  The hyperplane for ground-object contact is in red, while the robot-object contact hyperplane is in blue.  Each initial condition has its associated MPC cost.  In this example, the rightmost sample's LCS approximation allows the robot to most effectively foresee progressing the object toward the goal and correspondingly has the lowest sample cost.  Right:  These LCS approximations are well-defined even for more complicated geometries,
    only requiring witness points between the object and other collision geometry.}
    \label{fig:lcs_cartoons}
\end{figure*}

\subsection{Optimal Control through Contact}
\label{subsec:back:optimal}

We are interested in the MPC problem first posed in \cite{posa2014direct} for contact-rich systems with dynamics described by an NCP,
% \begin{subequations} \label{eq:mpc_nonlinear}
    \begin{align}
        \min_{x_k, u_k, \lambda_k} \; \sum_{k=0}^{N-1} \, &(x_k^T Q_k x_k + u_k^T R_k u_k)  + x_N^T Q_N x_N \label{eq:mpc_cost} \\
        \textrm{s.t.} \quad &x_{k+1} = g \left( x_k, u_k, \lambda_k \right), \label{eq:mpc_dynamics} \\
        & 0 \leq \lambda_k \perp \pi \left( x_k, u_k, \lambda_k \right) \geq 0, \label{eq:mpc_contact} \\
        & x_0 = x_\text{init}, \, (x_k, u_k) \in \mathcal{C}. \label{eq:mpc_constraints}
        % & \text{for} \; k \in \{ 0, \ldots, N \}.
    \end{align}
% \end{subequations}
This optimization problem solves for state $x_{0:N}$, control input $u_{0:N-1}$, and contact force $\lambda_{0:N-1}$ trajectories that satisfy the system's dynamics \eqref{eq:mpc_dynamics}, contact constraints \eqref{eq:mpc_contact}, and initial/other constraints \eqref{eq:mpc_constraints}, e.g. input limits or safety constraints.

Solving this MPC problem for even simple multi-contact systems is intractable at real-time rates due to the complicated nature of the discontinuous contact dynamics \eqref{eq:mpc_contact} and their impact on system dynamics \eqref{eq:mpc_dynamics}.  An alternative is to approximate them with expressions that are faster to evaluate.

\subsection{Consensus Complementarity Control (C3)}
\label{subsec:back:c3}

To simplify \eqref{eq:mpc_cost}-\eqref{eq:mpc_constraints}, we pose the control problem with the dynamics expressed as an LCS (substituting \eqref{eq:lcs} for \eqref{eq:mpc_dynamics} and \eqref{eq:mpc_contact}) using $x_\text{init}$ (and some nominal $u_0$, $\lambda_0$) as the LCS linearization point.
This is still computationally intensive because of the complementarity constraint \eqref{eq:lcs_contact}.  Consensus Complementarity Control (C3) \cite{aydinoglu2024consensus}, without guarantees of convergence, can approximate this optimization problem quickly.
C3 uses the alternating direction method of multipliers (ADMM) to iteratively solve the problem while satisfying only the dynamics \eqref{eq:lcs_dynamics} then only the contact constraints \eqref{eq:lcs_contact}.  This first stage is a quadratic program (QP) with linear equality and inequality constraints.  The second stage can be parallelized across time steps, since the contact constraints \eqref{eq:lcs_contact} depend only on step $k$.  Thus, the second stage can be formulated as a small-scale, sufficiently fast mixed-integer QP (MIQP).  With extra cost terms to encourage consensus between these two stages, C3 can converge to the solution to the optimization problem.
In practice, using a suboptimal solution at a faster rate is more performant than a more optimal solution at a slower rate, so it is beneficial to terminate C3 after a few ADMM iterations.

%%%%%%%%%%%%%%%%%%%%%%%%%%%%%%%
\section{Methods}
\label{sec:methods}

We extend the capabilities of local control by injecting global insights into the problem.
Using an LCS model as in C3 \cite{aydinoglu2024consensus} has negative consequences that motivate our approach 
(\S \ref{subsec:lcs_limitations}).  The key is to split the problem into an initial contact-free mode and a subsequent contact-rich mode (\S \ref{sec:methods:separate}).
To approximate the new bilevel optimization problem in real time, we sample low-dimensional end effector locations at the mode switch (\S \ref{sec:methods:sample}).  In closed-loop, our controller autonomously switches between modes based on cost and progress (\S \ref{sec:methods:switch}).  While our motivation and demonstrations use C3 \cite{aydinoglu2024consensus} as the local contact-rich controller, our hierarchical approach could wrap around other contact-rich controllers, e.g. \cite{cleac2021fast}.

\subsection{Limitations of LCS as an MPC Modeling Choice}
\label{subsec:lcs_limitations}

The LCS approximation of a system (as in \cite{aydinoglu2024consensus, cleac2021fast}) creates hyperplanes in $(x, u, \lambda)$ space with respect to every contact pair of interest.
While not strictly accurate\footnote{Precisely, the linearization of the gap function $\phi$ is a combination of the hyperplanes in the figure, with additional terms derived from the linearizations of the single-step dynamics.}, one can conceptualize these hyperplanes as approximating the non-linear geometry of an object via witness point tangent planes in $\mathbb{R}^3$ (Fig. \ref{fig:lcs_cartoons}).
With these visuals, it is evident the half-space boundaries can easily lock the robot into regions where it cannot apply positive normal force to move the object towards its goal (e.g. third from left in Fig. \ref{fig:lcs_cartoons}).  Friction cone constraints prevent planning negative normal forces, so the robot will plan to do nothing.

Thus, any LCS- or linearization-based controller is fundamentally local.  Past C3 demonstrations \cite{yang2024dynamic, aydinoglu2024consensus} required careful experimental design and/or additional heuristics to encourage or force the end effector out of regions where the local LCS view is antagonistically simplified.  These heuristics depend on the system and the goal, and bake in prior knowledge for what regions better enable task progress.  We seek a simple way to instill global knowledge into a controller that is locally effective.

\subsection{Separate Contact-Free and Contact-Rich Modes}
\label{sec:methods:separate}

First, consider that any optimal multi-contact control problem can be decomposed into an initial contact-free mode and a subsequent contact-rich mode.  The contact-free mode has no collisions between the robot and manipulated objects, and the contact-rich mode can make and break contact as it pleases.  This suggests a new bilevel optimization problem\footnote{Ignoring other constraints for brevity, including dynamics, contact dynamics, initial condition, input limits, etc.} given by
\begin{align} \label{eq:bilevel}
    \min_{K, x_\text{switch}} \left( \begin{matrix} \underset{x_k, u_k}{\min} \, \underset{k=0}{\overset{K}{\sum}} c_\text{free}(x_k, u_k) \\ \text{s.t. } x_K = x_\text{switch} \end{matrix} + \begin{matrix} \underset{x_k, u_k}{\min} \, \underset{k=K}{\overset{N}{\sum}} c_\text{rich}(x_k, u_k) \\ \text{s.t. } x_K = x_\text{switch} \end{matrix} \right),
\end{align}
whose outer problem selects a time step $K$ and system state $x_\text{switch}$ for \textit{when and where} the mode switches from contact-free to contact-rich, and whose inner problem minimizes the contact-free and contact-rich costs ($c_\text{free}$ and $c_\text{rich}$, respectively) given the intermediate state $x_\text{switch}$.  This loses no generality from the original contact-rich control problem \eqref{eq:mpc_cost}-\eqref{eq:mpc_constraints}.  To enable computational efficiency, we approximate in \eqref{eq:bilevel}:
\begin{enumerate}
    \item the contact-free mode as end effector repositioning (kinematic path planning);
    \item the contact-rich mode via local CI-MPC (C3 \cite{aydinoglu2024consensus}).
\end{enumerate}
The first approximation can be justified for quasi-static systems, where the object remains stationary during contact-free motion.  While no tasks are truly quasi-static, this is a common assumption in manipulation \cite{halm2024set, dogar2012planning, lynch1996stable}. As we will see in \S \ref{sec:experiments}, even for more dynamic tasks, this is not particularly limiting.  This means the effective output of the first stage is a new configuration for the robot collision geometry.  This configuration space can be tractable for optimization or search if we use a low-dimensional representation, such as a 3D end effector location (as used in our demonstrations).  Specifically, if a full system state $x$ contains $[ q_\text{ee}, q_\text{obj}, v ]$, we impose $x_\text{switch} = [ q_{\text{ee},\text{switch}}, q_{\text{obj},\text{init}}, v_\text{init}]$ where only $q_{\text{ee},\text{switch}}$ differs from $x_\text{init}$.  Additionally, the cost of the contact-free stage can be penalized by a simple end effector travel cost.

The second simplification is based on the efficacy of C3 for the more challenging portion of the problem, with LCS dynamics (including hyperplane definitions, as in Fig. \ref{fig:lcs_cartoons}) local to the starting state with $x_\text{switch}$.  By making global decisions via contact-free path planning and sequencing with contact-rich motions, we are able to mitigate the limitations inherent in the locality of C3's contact-implicit control.

\subsection{Sample Switching States}
\label{sec:methods:sample}

Even with simple travel cost and C3 approximations to the contact-free and contact-rich modes, respectively, the bilevel optimization problem is intractable to solve in real time because it would require solving C3 for every candidate $x_\text{switch}$.  Instead, we sample a small set of $x_\text{switch}$, which crucially can be evaluated in parallel.
This sample set always includes the current system state, i.e. our hierarchical controller always knows the contact-rich plan and can enter/remain in the contact-rich mode whenever it wants.
If the controller is currently in contact-free mode, the sample set additionally includes the actively pursued $q_{\text{ee,switch}}$ (called $q_\text{ee,switch,prev}$ coming from the previous loop's $x_\text{switch}$ i.e. $x_\text{switch,prev}$) to continually update its sample cost.
At every control loop, our controller computes the C3 costs for the $x_\text{switch}$ candidates, adding the contact-free travel costs appropriately.  We assume access to a sampling strategy $\texttt{SampleEE}(x_\text{init}, x_\text{switch,prev}, n_s)$ which draws $n_s$ $x_\text{switch}$ samples for potential end effector locations.  The first sample is always $q_\text{ee,init}$, and the second is $q_\text{ee,switch,prev}$, if there is one.
To evaluate costs, we assume access to the C3 algorithm and the function $\texttt{C3Cost}(x_\text{sample}, x_\text{goal})$.
This sampling and evaluation step is detailed by Algorithm \ref{alg:sample_eval}.

\begin{algorithm}
    \caption{Sample and Evaluate} \label{alg:sample_eval}
    \begin{algorithmic}[1]
        \REQUIRE $x_\text{init}, x_\text{goal}, x_\text{switch,prev}$ if there was one, number of samples $n_s$, sampling strategy \texttt{SampleEE}, travel cost weight $w_\text{travel}$, C3 algorithm and associated cost \texttt{C3Cost}

        \STATE $q_{\text{sample}, \text{ee}, 1:n_s} \gets \texttt{SampleEE}(x_\text{init}, x_\text{switch,prev}, n_s)$
        
        \textbf{\\Parallelizing samples, solve C3 and impose travel cost.}
        \vspace{-1em}
        \FOR {$i \in \{1, \dots, n_s \}$}
            \STATE $c_\text{travel} \gets w_\text{travel} \norm{q_\text{ee,init} - q_{\text{ee,sample},i}}$
            \STATE $x_{\text{sample}, i} \gets [q_{\text{ee}, \text{sample}, i}, q_{\text{obj}, \text{init}}, v_\text{init}]$
            \STATE $c_{\text{sample}, i} \gets \texttt{C3Cost}(x_{\text{sample}, i}, x_\text{goal}) + c_\text{travel}$ \label{alg:line_parallel}
        \ENDFOR

        \RETURN $(x_\text{sample}, c_\text{sample})_{1:n_s}$
    \end{algorithmic}
\end{algorithm}

\subsection{Switch Modes Based on Costs and Progress}
\label{sec:methods:switch}

Given a set of $x_\text{switch}$ candidates and associated inner costs $c_\text{sample}$, we must pursue one.  If we pick the current state, our controller executes the actions computed during the C3 solve.
If we pick a different state, our controller commands a collision-free path to the new end effector location.
A natural selection algorithm might be to choose the sample with the lowest associated cost.  However, we find this results in indecisive behavior. We attribute this primarily to C3 variability, as C3 is a local algorithm with no convergence guarantees.  To mitigate these effects, we add hysteresis on cost comparisons to keep the controller in the current mode, barring a clearly better alternative. We define hysteresis values $h_\text{free-to-rich}$, $h_\text{rich-to-free}$, and $h_\text{free-to-free}$ (which relates to switching between different relocation targets within the contact-free mode). We choose $h_\text{rich-to-free}$ to be relatively large, so that the algorithm will not mistakenly abandon a contact-rich mode that is progressing toward the goal. However, this alone creates the risk of the controller becoming stuck while no progress is made. To address this, we force a transition to the contact-free mode if the manipulated object fails over a specified period of time to make progress toward the goal. When this occurs, the controller pursues the sampled $x_\text{sample}$ with the lowest cost.

To increase the likelihood of pursuing a high-quality $x_\text{sample}$ after leaving contact-rich mode, we maintain a buffer of relevant sampled end effector positions and their associated costs.  The sample buffer is pruned of samples whose associated object locations at the time of computing cost are too far from the current object location.
At every control loop, we:
\begin{enumerate}
    \item Update $x_\text{init}$ from proprioception/vision-based sensing.
    \item Generate and evaluate samples via Algorithm \ref{alg:sample_eval}.
    \item Maintain the sample buffer:  remove outdated samples based on object movement and add those from Step 2.
    \item Determine the controller mode. Select the lowest cost sample from the buffer, accounting for hysteresis which privileges the current mode. If no progress has been made, force a switch to contact-free mode.
    \item Execute a plan based on the current mode.  If in contact-rich mode, execute the plan from solving C3 with $x_\text{init}$.  If in contact-free mode, follow a collision-free path from $x_\text{init}$ to the pursued sample.
\end{enumerate}
We repeat this with receding horizon, enabling the control to adapt to observed dynamics, considering more samples with every control loop.  Fig. \ref{fig:algorithm} depicts one control loop.

%%%%%%%%%%%%%%%%%%%%%%%%%%%%%%%
\section{Experiments}
\label{sec:experiments}

\subsection{Task, State Representation, and Contact Modeling}
\begin{figure}[t]
    \centering
    \includegraphics[width=\linewidth,trim={0mm 0mm 0mm 0mm},clip]{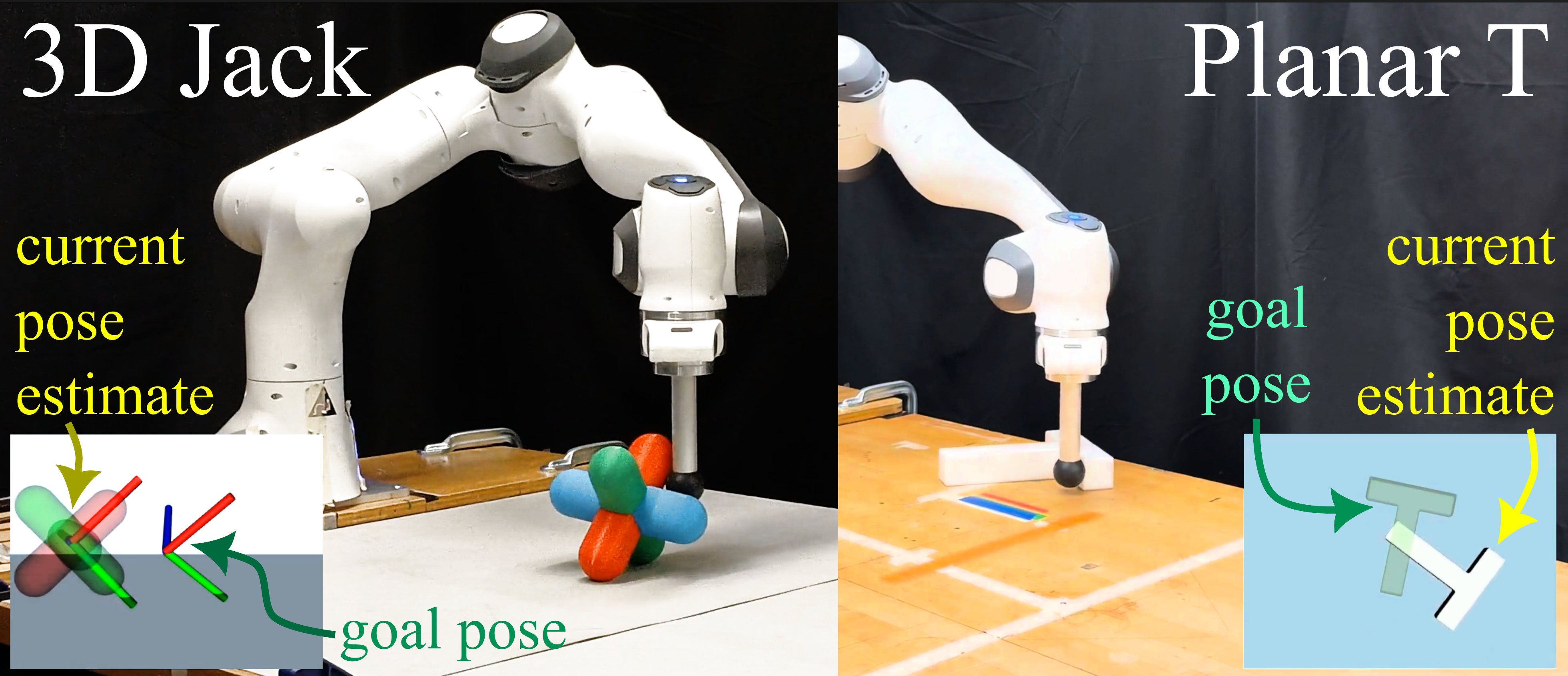}
    \vspace{-2em}
    \caption{We demonstrate our controller on two manipulation examples:  3D pose goals with a jack (left), and 2D planar pose goals with a T (right). The inset renderings in the bottom corners depict a view of the object's goal pose relative to the current pose estimate.}
    \label{fig:experiments}
\end{figure}

To validate our controller, we test with a Franka Panda arm equipped with a spherical end effector to manipulate an object to a pose goal.  When the object goal is reached within position and rotation tolerances, a new pose goal is randomly generated, demonstrating generalization over initial and goal poses.  We test on two hardware examples (3D jack and planar T, both shown in Fig. \ref{fig:experiments}) and on our 3D jack example in simulation for more direct comparison to baselines (more detail in \S \ref{subsec:comparisons}).
Our controller operates on LCS dynamics
where $x \in \mathbb{R}^{19}$ (end effector position, object position, object orientation, and their velocities), $u \in \mathbb{R}^3$ (cartesian forces applied to the end effector), and $\lambda \in \mathbb{R}^{16}$ (4 contact pairs with a 4-sided polyhedral friction cone approximation \cite{anitescu1997formulating}).

\begin{figure}
    \centering
    \includegraphics[width=0.5\linewidth,trim={40mm, 18mm, 110mm, 28mm},clip]{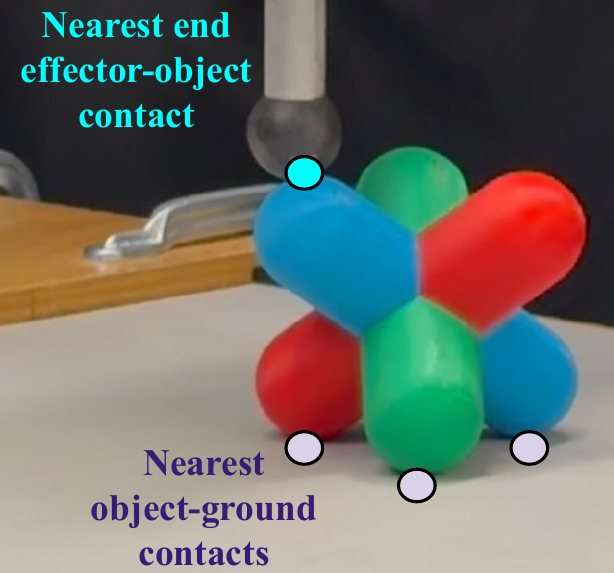}
    \vspace{-0.5em}
    \caption{Illustration of contact points considered for the jack object.}
    \label{fig:jack_contacts}
\end{figure}

Our approach requires manual enumeration of contact \textit{geometries}, but uses collision detection between them to identify possible contact \textit{points} at every control loop.  These points automatically change with motion, and the approach can accommodate any geometries handled by the collision detection algorithm.  For both examples, one contact pair uses the closest witness point on the object to the end effector, and the remaining 3 contact pairs are for object-ground contacts.  For the jack, we obtain the closest point per capsule to the ground.  A visualization of all the jack contact points for a particular configuration is illustrated in Fig. \ref{fig:jack_contacts}.  For the T, we define small spheres at the 3 distal ends as witnesses to the ground.  For this planar example, we use the full 3D states of the end effector and T but add a constraint in the contact-rich stage (C3) to enforce the end effector maintains a pre-defined height.

\subsection{Implementation}
\begin{figure}
    \centering
    \includegraphics[width=\linewidth]{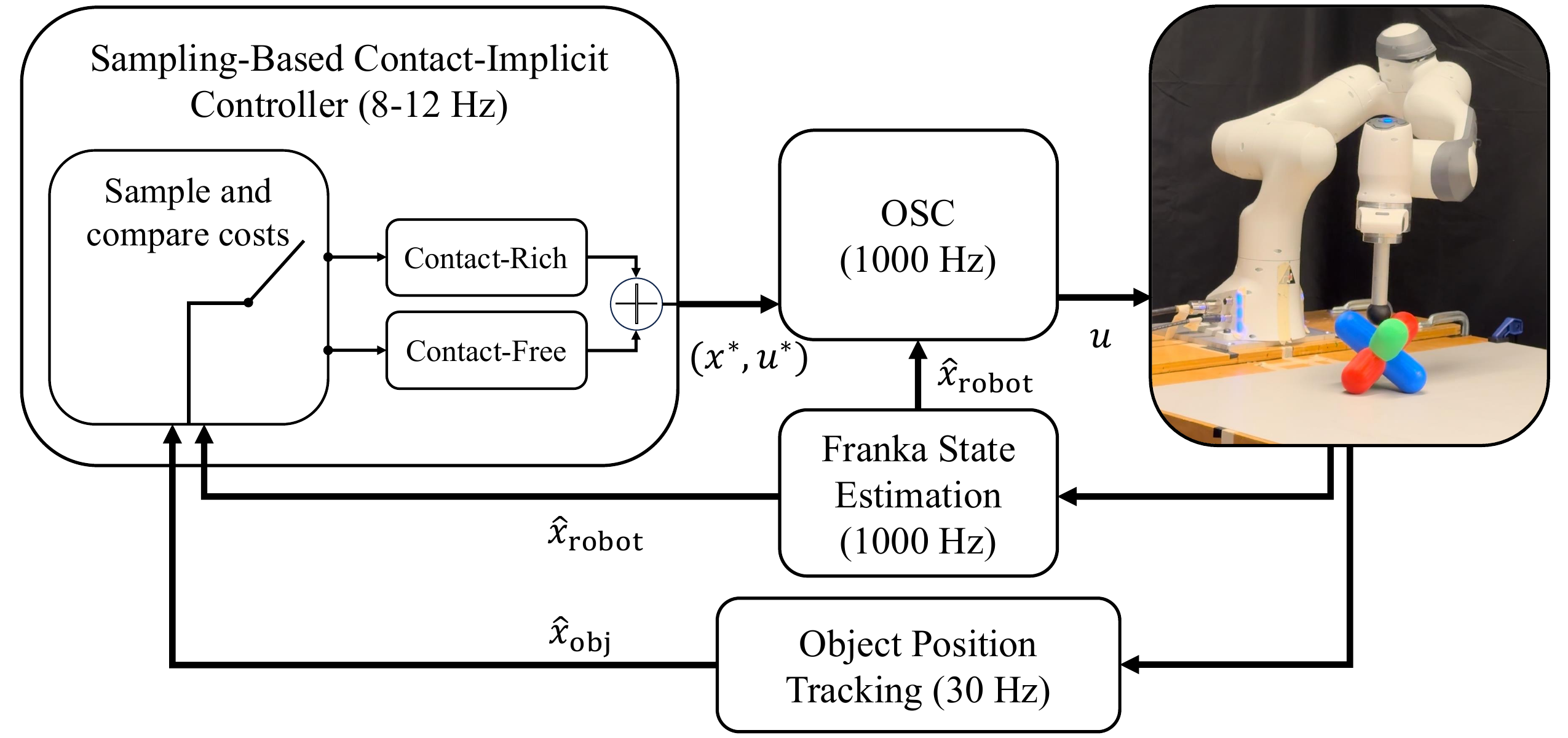}
    \vspace{-2em}
    \caption{A block diagram depicts our hierarchical controller (top left block), which performs real-time CI-MPC globally by sampling potential end effector locations for switching from contact-free to contact-rich mode.}
    \label{fig:architecture}
\end{figure}

Our controller and accompanying diagram (Fig. \ref{fig:architecture}) is implemented in C++ in Drake \cite{tedrake2019drake}.  As in \cite{yang2024dynamic}, an operational space controller (OSC) \cite{wensing2013generation}
tracks task-space commands specified by our policy at 8-12Hz, via joint-level control at 1kHz.
FoundationPose \cite{wen2024foundationpose} estimates object state and runs at 30Hz with a D455 RealSense RGBD camera.
Our setup uses two computers:  a 13th generation Intel Core i9-13900KF with 32 threads (for our controller) and an NVIDIA GeForce RTX 4090 GPU (for FoundationPose), and an Intel i7-8700K processor (for our OSC and robot drivers) equipped with a real-time kernel for communicating with the Franka.  Inter-computer communication occurs over LCM \cite{huang2010lcm}.

\subsection{Sampling Parameters}
We consider three samples (including the end effector's current location) in parallel with every control loop ($n_s = 3$).
We solve C3's QP with OSQP \cite{stellato2020osqp} and MIQP with Gurobi \cite{gurobi}.
Our samples and the C3 MIQP both utilize parallelization, so we use the maximum number of threads in our hardware setup.  Because of this, additional samples slow down the control rate and decrease performance.  With this setup, our controller runs at 8-12Hz for all experiments.  We randomly sample end effector locations on a sphere around the jack and on an enlarged planar perimeter for the T.  While more sophisticated sampling is compatible with our approach, these simple distributions are notably effective.

\subsection{Cost Parameters}
To avoid aggressive behavior from high costs, we truncate the final goal's displacement from the current object location (15cm, 2 radians).  This bounds the errors encountered by the MPC and mitigates indecision when orientation error approaches $\pi$ radians, at which any rotation direction is equally effective.  This orientation truncation requires hysteresis on the rotation axis to ensure stability of the direction of rotation.

For the $Q$ cost matrix, a typical diagonal structure is sufficient for position and velocity errors in our experiments.  However, orientation presents some challenges.  The true orientation error we desire to minimize is $\theta_\text{error}^2$, where $\theta_\text{error}$ is the scalar angle of the relative rotation between the current and goal orientations.
From quaternions, this is calculated as
\begin{subequations}
    \begin{align}
        \theta_\text{error} &= \left(\arctan\left( \frac{\norm{q_{\text{rel},x}^2 + q_{\text{rel},y}^2 + q_{\text{rel},z}^2}}{q_{\text{rel},w}} \right)\right)^2, \\
        \text{where} \quad q_\text{rel} &= q_\text{quat,curr}^{-1} \otimes q_\text{quat,goal},
    \end{align}
\end{subequations}
for $\otimes$ as quaternion product.  The $\arctan$ indicates a problematic region where its argument is zero -- this occurs precisely when $q_\text{quat,curr} = q_\text{quat,goal}$.  The landscape is not strictly convex at this point and is locally non-convex.  Thus, the naive 2-norm error between the elements of $q_\text{quat,curr}$ and $q_\text{quat,goal}$,
\begin{align}
    \tilde{\theta}_\text{error}^2 &= \norm{q_\text{quat,curr} - q_\text{quat,goal}}^2.
\end{align}
poorly captures the true $\theta_\text{error}$ when it is small.  To address this, we set the 4x4 object quaternion portion of $Q$ (throughout the entire MPC horizon) to be the Hessian of $\theta_\text{error}^2$ with respect to the elements of the current quaternion, about the $q_\text{quat,curr}, q_\text{quat,goal}$ operating point.
We regularize this Hessian, adding $| \gamma | \cdot \mathbb{I}_{4\times4}$, where $\gamma$ is its most negative eigenvalue to ensure positive-semi definiteness.  Implementing this portion of $Q$ is a critical step to effectively and reliably track orientation.

\subsection{Comparisons}
\label{subsec:comparisons}
Due to the inability to escape geometric local minima, C3 fails essentially 100\% of the time on our tasks and thus is not compared.  We compare with two ablations.  Naive-R \underline{\textbf{r}}andomly selects an $x_\text{sample}$ when the controller is unproductive; in contrast, our controller selects the $x_\text{sample}$ with the lowest associated CI-MPC cost and can switch modes due to progress or cost.  Naive-B only considers the $x_\text{sample}$ \underline{\textbf{b}}ehind the object from the goal, testing whether a trivial region of space around the object is sufficient or if broader exploration is required.

Additionally, we compare with MuJoCo MPC (MJPC) with predictive sampling \cite{howell2022predictive} on the 3D jack task in simulation.  As with our controller, we use MJPC as an online planner operating on a reduced model, abstracting the end effector as controlled in $xyz$ only, whose motion is tracked via our joint-level OSC simulated in Drake.  Unlike our controller, the MJPC planner models the nonlinear dynamics of the floating end effector moving in 3D with the jack and environment.  
Any comparison is sensitive to tuning, and we put forth a best-faith effort to tune MJPC.
We found tighter control input limits to be helpful in preventing wild motions, selected a noise parameter of 0.295 to balance sample exploration and previous plan exploitation, and tuned state and input costs.
MJPC performed best with control input splines of 5 knot points over a predictive horizon of 0.8 seconds, reasonably longer than our policy's 0.25 seconds, since we effectively get longer-term insight via our contact-free sampling.

%%%%%%%%%%%%%%%%%%%%%%%%%%%%%%%
\section{Results}
\label{sec:results}

\begin{figure}[t]
    \centering
    3D Jack Hardware Goals Achieved within Time Limit
    \includegraphics[width=0.8\linewidth,trim={0mm, 0mm, 0mm, 8mm},clip]{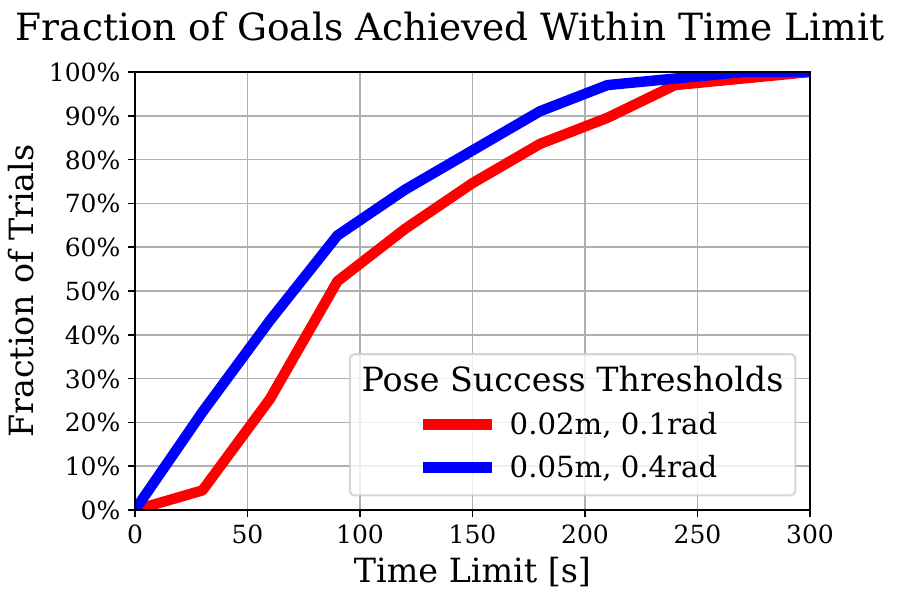}
    \vspace{-0.5em}
    \caption{Cumulative distribution for time to goal, using sets of tight and loose position and orientation tolerances.}
    \label{fig:cdf}
\end{figure}

\begin{figure}[t]
    \centering
    \includegraphics[width=\linewidth,trim={0mm 0mm 0mm 0mm},clip]{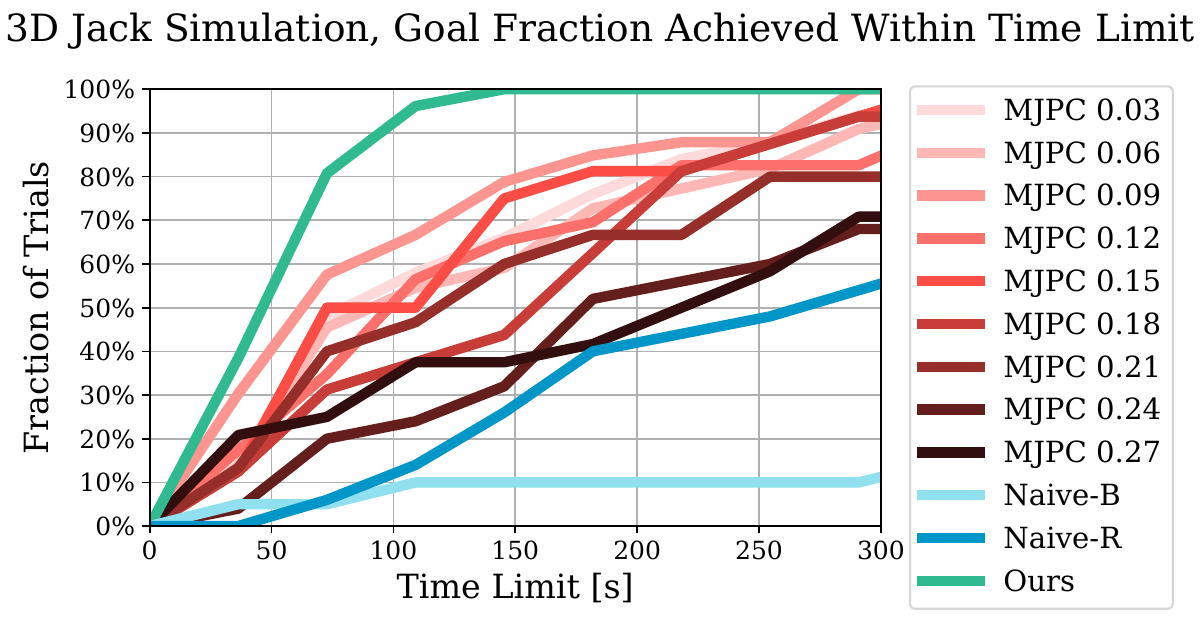}
    \includegraphics[width=\linewidth,trim={0mm 0mm 0mm 0mm},clip]{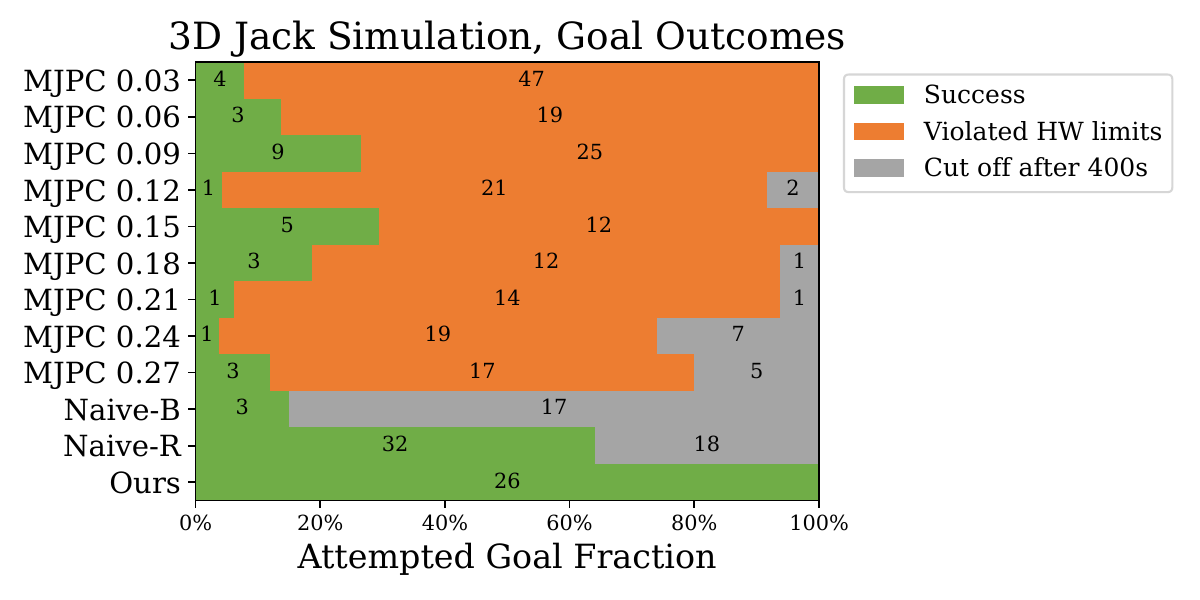}
    \vspace*{-20ex}  % Tune this to the image height.
    \begin{flushleft}
    \scriptsize{\qquad \qquad \qquad \qquad \qquad \qquad \qquad \qquad \qquad \qquad \qquad \qquad \qquad Trial counts \par \qquad \qquad \qquad \qquad \qquad \qquad \qquad \qquad \qquad \qquad \qquad \qquad \qquad annotated on chart.}
    \end{flushleft}
    \vspace*{20ex}
    \vspace{-5em}
    \caption{Our approach outperforms all other baselines %MJPC
    in 3D jack manipulation in simulation with high precision tolerances (2cm and 0.1 radians pose error).  Our controller achieves goals faster than MJPC (top) while also avoiding hardware limit violations (bottom) such as joint velocity, joint torque, and workspace limits.
    The MJPC lines are annotated with the values of the end effector velocity cost weight, showing decreased performance but ineffective ability to reduce hardware violations at higher end effector velocity costs.  Ablations of our approach (Naive-R and Naive-B) avoid hardware limit violations (bottom) but are substantially less performant in terms of success rate and time-to-goal (top).}
    \label{fig:mjpc_cdf}
\end{figure}

See our \href{https://approximating-global-ci-mpc.github.io/}{project webpage} for extended results.
We evaluate performance under two sets of success thresholds: the tight set requires under 2cm and 0.1 radians (5.7$^\circ$) of error, and coarse set under 5cm and 0.4 radians (22.9$^\circ$). Related works (e.g. \cite{li2024drop, chen2023visual}) often use this coarse threshold.
For all experiments, we execute the controller until the tight threshold is realized, then switch to the next goal pose. In post-processing, we back-compute the time-to-goal under coarse thresholds (Table \ref{tab:time_to_goal_table}).

\begin{table}[t]
\centering
\resizebox{\columnwidth}{!}{
\begin{tabular}{cccc}
\toprule
Mean $\pm \sigma$ & \multicolumn{2}{c}{Time to Goal (s) within Pose Tolerances} \\ %\cline{2-3}
$[$Min, Max$]$ & Tight:  2cm, 0.1rad & Loose:  5cm, 0.4rad \\

\midrule
\midrule
\textbf{HW 3D Jack (Ours)} & $109.20 \pm 64.24$ & $84.86 \pm 60.54$ \\
\textbf{67 success (0 HWV), 0 fail} & $[17.40, 292.07]$ & $[5.20, 257.74]$ \\
\midrule
\textbf{Sim 3D Jack (Ours)} & $49.31 \pm 30.35$ & $33.84 \pm 26.39$ \\
\textbf{26 success (0 HWV), 0 fail} & $[9.71, 124.70]$ & $[7.97, 97.82]$ \\
\midrule
\textbf{Sim 3D Jack (Naive-R)} & $182.34 \pm 84.66$ & $138.17 \pm 78.89$ \\
\textbf{32 success (0 HWV), 18 fail} & $[63.52, 352.75]$ & $[18.64, 297.00]$ \\
\midrule
\textbf{Sim 3D Jack (Naive-B)} & $140.86 \pm 121.46$ & $49.67 \pm 32.25$ \\
\textbf{3 success (0 HWV), 17 fail} & $[18.55, 306.47]$ & $[17.69, 93.82]$ \\
\midrule
\textbf{Sim 3D Jack (MJPC)} & $ 107.91 \pm 112.38$ & $68.00 \pm 83.50$ \\
\textbf{34 success (25 HWV), 0 fail} & $[3.30, 567.69]$ & $[1.51, 343.79]$ \\
\midrule
\midrule
\textbf{HW Planar Push-T (Ours)} & $30.45 \pm 13.11$ & $17.43 \pm 7.59$ \\
\textbf{106 success (0 HWV), 0 fail} & $[7.50, 79.43]$ & $[3.86, 42.00]$ \\
\bottomrule
\end{tabular}
}
\vspace{0.3em}
\caption{Performance metrics of hardware (HW) and simulation (Sim) experiments under tight and loose tolerances.  HW limit violations (HWV) indicated.  MJPC results use highest-performing end effector velocity cost of 0.09.}
\label{tab:time_to_goal_table}
\end{table}

\subsection{3D Jack}
\label{subsec:results:jack}

\begin{figure}[t]
    \centering
    3D Jack Hardware Manipulation Examples
    \includegraphics[width=\linewidth,trim={0mm 0mm 190mm 12mm},clip]{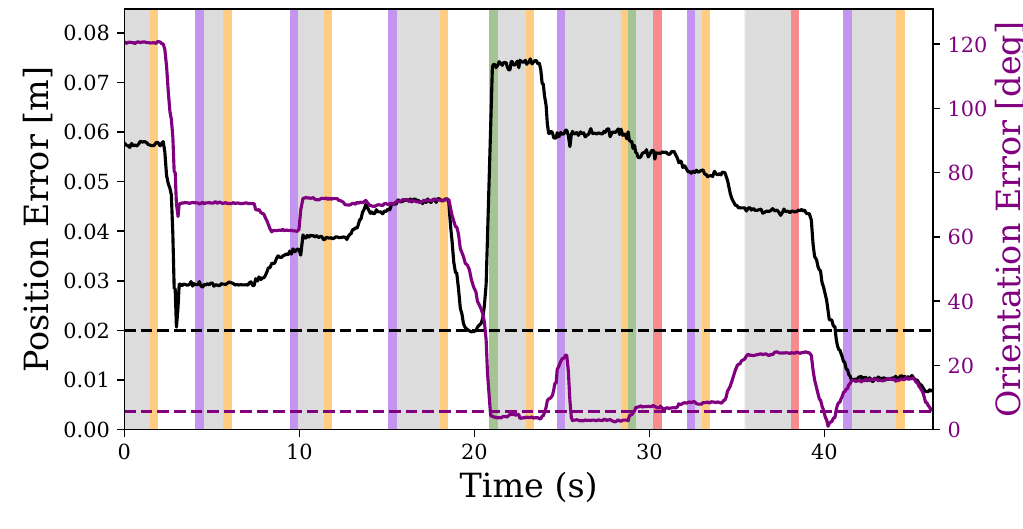}
    \\
    \vspace{-0.5em}
    \includegraphics[width=\linewidth,trim={0mm 0mm 190mm 12mm},clip]{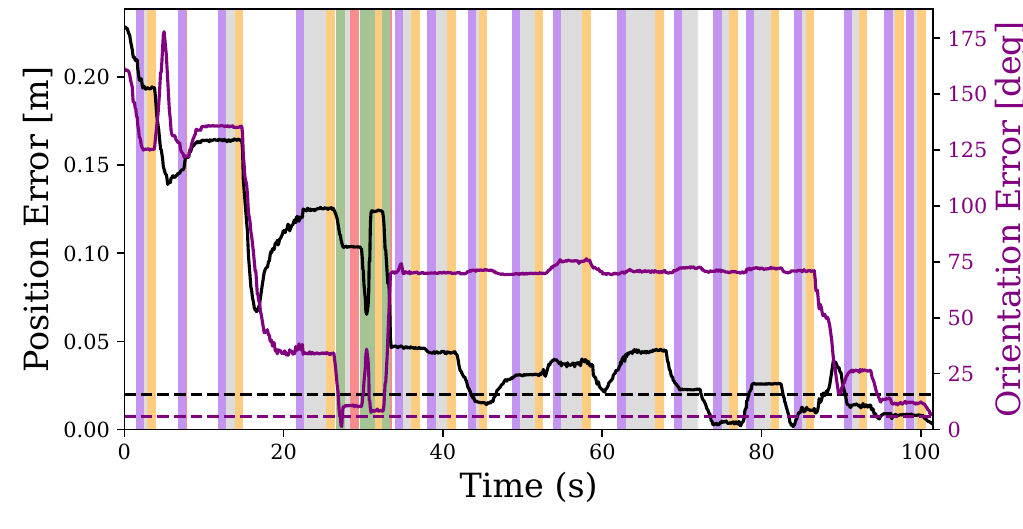}
    \\
    \vspace{-0.5em}
    \includegraphics[width=0.7\linewidth,trim={180mm 10mm 32mm 15mm},clip]{Figures/jack_results/shading_goal_7.pdf}
    \vspace{-0.5em}
    \caption{Jack position and orientation errors over time, with contact-free (grey) and contact-rich (white) modes shaded and mode switching reasons labeled.}
    \label{fig:mode_switching}
\end{figure}

Fig. \ref{fig:cdf} combines hardware results from four continuous experiments of 21, 16, 15, and 15 achieved random pose goals. All experiments terminated when the robot hit workspace safety limits.
The time-to-goal cumulative distribution shows that while most goals are reached relatively quickly, there is a long tail with some challenging targets requiring more time.

Fig. \ref{fig:mode_switching} depicts two single-goal trials achieved on hardware.
These show the controller's balancing position and rotation progress, occasionally sacrificing one to make progress on the other.
Both trials demonstrate cost- and progress-based mode transitions.  Most often, errors do not change during contact-free mode, since the object usually stays fixed while the robot relocates.  One cause of inefficiency is simultaneous low position error and high rotation error.  The only terminal failure condition is when the object is pushed to the boundary of the safe workspace, causing the robot to cross our workspace safety limits.  In all other trials, our controller's persistence eventually brings the object to the goal in every test.

\subsubsection{Sim-to-Real Gap and Comparisons}
Our controller, unsurprisingly, achieves pose goals faster in simulation than on hardware (see 49.31s simulation average compared to 109.20s hardware average in Table \ref{tab:time_to_goal_table}).  This gap, common in the literature, can be partly explained by state estimate errors, incorrect models, and FoundationPose's added computational load.  We reiterate our approach is not optimal; it is a real-time CI-MPC approach that makes all its decisions on-the-fly.  While not optimal, our controller is demonstrably effective on difficult tasks and outperforms MJPC \cite{howell2022predictive} and our ablations in simulation on our 3D jack example (Table \ref{tab:time_to_goal_table}, Fig. \ref{fig:mjpc_cdf}).
Further, our controller and its ablation variants (Naive-R and Naive-B) satisfy Franka hardware limits (joint velocity/torque, workspace limits), while MJPC nearly always violates at least one.  Given this high hardware limit violation (HWV) rate, we did not feel safe deploying MJPC on the real robot.  While impossible to compare against all possible MJPC weights, we investigate the role of end effector velocity cost weight, due to its role in the controller's speed.  Increasing this cost weight makes MJPC less performant, yet does not prevent HWV.

For the majority of trials, Naive-B is incapable of bringing the object to the goal.  Its failure mode is when the sample behind the goal is unhelpful, so no contact is established during the contact-rich mode.  Thus the sample does not update in response to object movement, and the robot will continually switch between contact-free and contact-rich modes without further moving the object.  In contrast, Naive-R (like our approach) should always bring the object to the goal, though requiring more time to do so, indicating using the CI-MPC cost is a useful identifier of more effective contact-rich modes.

\subsection{Planar T Pushing}
\label{subsec:results:t}
The statistics in Table \ref{tab:time_to_goal_table} for the hardware planar T experiments combine four continuous experiments of 56, 20, 20, and 10 successfully achieved random planar pose goals in a row.  With an average of 30.45s to achieve high-precision pose goals, our approach achieves a state-of-the-art time-to-goal competitive with other works on this example.  Notably, while some prior works with this example use data-driven approaches (e.g. pre-training with Diffusion Policy \cite{chi2023diffusion}) or require offline computation (e.g. offline trajectory optimization \cite{kang2025global}), our approach demonstrates only the object model is required, and generalization to different goal poses is a natural byproduct absent from these prior works.

%%%%%%%%%%%%%%%%%%%%%%%%%%%%%%%
\section{Limitations}
\label{sec:limitations}

While we contribute a solution to the global, 3D manipulation problem, our controller still took $\sim$1.8 minutes on average over 67 trials to achieve precise SE(3) goals. 
We reiterate the generality of our approach and acknowledge these tasks are challenging.  However, addressing inefficiencies is future work.
Like all model-based methods, our controller requires dynamics models of the robot, objects, and environment, preventing use in truly novel scenarios.
Our demonstrations used one spherical end effector, which we reasonably modeled as a single robot-object contact pair.  If applied to a more dexterous manipulator such as a multi-fingered hand, contact pairs and state size would increase, slowing down control rates.

%%%%%%%%%%%%%%%%%%%%%%%%%%%%%%%
\section{Conclusion} 
\label{sec:conclusion}

Our model-based controller performs generalizable, precise pose-driven tasks through multi-contact dynamics.  By splitting the control problem into contact-free and contact-rich stages, we reap the benefits of global exploration when we sample new end effector locations plus local efficacy when existing CI-MPC methods take over after arriving at a desired location.  In closed loop, the result is a persistent, globally-aware controller that can robustly reach precise pose goals without intervention.

\bibliographystyle{IEEEtran}
\bibliography{references}

@article{toussaint2018differentiable,
  title={Differentiable physics and stable modes for tool-use and manipulation planning},
  author={Toussaint, Marc A and Allen, Kelsey Rebecca and Smith, Kevin A and Tenenbaum, Joshua B},
  year={2018},
  publisher={Robotics: Science and systems foundation}
}

@inproceedings{kamat2022bitkomo,
  title={Bitkomo: Combining sampling and optimization for fast convergence in optimal motion planning},
  author={Kamat, Jay and Ortiz-Haro, Joaquim and Toussaint, Marc and Pokorny, Florian T and Orthey, Andreas},
  booktitle={2022 IEEE/RSJ International Conference on Intelligent Robots and Systems (IROS)},
  pages={4492--4497},
  year={2022},
  organization={IEEE}
}

@inproceedings{chavan2019sampling,
  title={Sampling-based planning of in-hand manipulation with external pushes},
  author={Chavan-Dafle, Nikhil and Rodriguez, Alberto},
  booktitle={Robotics Research: The 18th International Symposium ISRR},
  pages={523--539},
  year={2019},
  organization={Springer}
}

@article{stellato2020osqp,
  title={OSQP: An operator splitting solver for quadratic programs},
  author={Stellato, Bartolomeo and Banjac, Goran and Goulart, Paul and Bemporad, Alberto and Boyd, Stephen},
  journal={Mathematical Programming Computation},
  volume={12},
  number={4},
  pages={637--672},
  year={2020},
  publisher={Springer}
}

@misc{gurobi,
  author = {{Gurobi Optimization, LLC}},
  title = {{Gurobi Optimizer Reference Manual}},
  year = 2024,
  url = "https://www.gurobi.com"
}

@book{stewart2011dynamics,
  title={Dynamics with Inequalities: impacts and hard constraints},
  author={Stewart, David E},
  year={2011},
  publisher={SIAM}
}

@article{tedrake2019drake,
  title={Drake: Model-based design and verification for robotics},
  author={Tedrake, R and others},
  year={2019}
}

@inproceedings{huang2010lcm,
  title={LCM: Lightweight communications and marshalling},
  author={Huang, Albert S and Olson, Edwin and Moore, David C},
  booktitle={2010 IEEE/RSJ International Conference on Intelligent Robots and Systems},
  pages={4057--4062},
  year={2010},
  organization={IEEE}
}

@article{kang2025global,
  title={Global Contact-Rich Planning with Sparsity-Rich Semidefinite Relaxations},
  author={Kang, Shucheng and Liu, Guorui and Yang, Heng},
  journal={arXiv preprint arXiv:2502.02829},
  year={2025}
}

@inproceedings{wensing2013generation,
  title={Generation of dynamic humanoid behaviors through task-space control with conic optimization},
  author={Wensing, Patrick M and Orin, David E},
  booktitle={2013 IEEE International Conference on Robotics and Automation},
  pages={3103--3109},
  year={2013},
  organization={IEEE}
}

@article{cheng2023enhancing,
  title={Enhancing Dexterity in Robotic Manipulation via Hierarchical Contact Exploration},
  author={Cheng, Xianyi and Patil, Sarvesh and Temel, Zeynep and Kroemer, Oliver and Mason, Matthew T},
  journal={arXiv preprint arXiv:2307.00383},
  year={2023}
}

@article{cleac2021fast,
  title={Fast contact-implicit model-predictive control},
  author={Cleac'h, Simon Le and Howell, Taylor and Schwager, Mac and Manchester, Zachary},
  journal={arXiv preprint arXiv:2107.05616},
  year={2021}
}

@inproceedings{aceituno2022hierarchical,
  title={A Hierarchical Framework for Long Horizon Planning of Object-Contact Trajectories},
  author={Aceituno, Bernardo and Rodriguez, Alberto},
  booktitle={2022 IEEE/RSJ International Conference on Intelligent Robots and Systems (IROS)},
  pages={189--196},
  year={2022},
  organization={IEEE}
}

@inproceedings{yang2024dynamic,
  title = {Dynamic On-Palm Manipulation via Controlled Sliding},
  author = {Yang, Will and Posa, Michael},
  year = {2024},
  month = jul,
  arxiv = {2405.08731},
  booktitle = {Robotics: Science and Systems (RSS)},
  website = {https://dynamic-controlled-sliding.github.io/},
  url = {https://roboticsconference.org/program/papers/12/}
}

@article{pang2023global,
  title={Global planning for contact-rich manipulation via local smoothing of quasi-dynamic contact models},
  author={Pang, Tao and Suh, HJ Terry and Yang, Lujie and Tedrake, Russ},
  journal={IEEE Transactions on Robotics},
  year={2023},
  publisher={IEEE}
}

@article{suh2022bundled,
  title={Bundled gradients through contact via randomized smoothing},
  author={Suh, Hyung Ju Terry and Pang, Tao and Tedrake, Russ},
  journal={IEEE Robotics and Automation Letters},
  volume={7},
  number={2},
  pages={4000--4007},
  year={2022},
  publisher={IEEE}
}

@inproceedings{wen2024foundationpose,
  title={Foundationpose: Unified 6d pose estimation and tracking of novel objects},
  author={Wen, Bowen and Yang, Wei and Kautz, Jan and Birchfield, Stan},
  booktitle={Proceedings of the IEEE/CVF Conference on Computer Vision and Pattern Recognition},
  pages={17868--17879},
  year={2024}
}

@inproceedings{antonova2023rethinking,
  title={Rethinking optimization with differentiable simulation from a global perspective},
  author={Antonova, Rika and Yang, Jingyun and Jatavallabhula, Krishna Murthy and Bohg, Jeannette},
  booktitle={Conference on Robot Learning},
  pages={276--286},
  year={2023},
  organization={PMLR}
}

@article{howell2022predictive,
  title={Predictive sampling: Real-time behaviour synthesis with mujoco},
  author={Howell, Taylor and Gileadi, Nimrod and Tunyasuvunakool, Saran and Zakka, Kevin and Erez, Tom and Tassa, Yuval},
  journal={arXiv preprint arXiv:2212.00541},
  year={2022}
}

@inproceedings{williams2016aggressive,
  title={Aggressive driving with model predictive path integral control},
  author={Williams, Grady and Drews, Paul and Goldfain, Brian and Rehg, James M and Theodorou, Evangelos A},
  booktitle={2016 IEEE International Conference on Robotics and Automation (ICRA)},
  pages={1433--1440},
  year={2016},
  organization={IEEE}
}

@article{halm2024set,
  title={Set-valued rigid-body dynamics for simultaneous, inelastic, frictional impacts},
  author={Halm, Mathew and Posa, Michael},
  journal={The International Journal of Robotics Research},
  pages={02783649241236860},
  year={2024},
  publisher={SAGE Publications Sage UK: London, England}
}

@article{dogar2012planning,
  title={A planning framework for non-prehensile manipulation under clutter and uncertainty},
  author={Dogar, Mehmet R and Srinivasa, Siddhartha S},
  journal={Autonomous Robots},
  volume={33},
  pages={217--236},
  year={2012},
  publisher={Springer}
}

@article{lynch1996stable,
  title={Stable pushing: Mechanics, controllability, and planning},
  author={Lynch, Kevin M and Mason, Matthew T},
  journal={The international journal of robotics research},
  volume={15},
  number={6},
  pages={533--556},
  year={1996},
  publisher={Sage Publications Sage CA: Thousand Oaks, CA}
}

@article{li2024drop,
  title={DROP: Dexterous Reorientation via Online Planning},
  author={Li, Albert H and Culbertson, Preston and Kurtz, Vince and Ames, Aaron D},
  journal={arXiv preprint arXiv:2409.14562},
  year={2024}
}

@article{chen2023visual,
  title={Visual dexterity: In-hand reorientation of novel and complex object shapes},
  author={Chen, Tao and Tippur, Megha and Wu, Siyang and Kumar, Vikash and Adelson, Edward and Agrawal, Pulkit},
  journal={Science Robotics},
  volume={8},
  number={84},
  pages={eadc9244},
  year={2023},
  publisher={American Association for the Advancement of Science}
}

@article{chi2023diffusion,
  title={Diffusion policy: Visuomotor policy learning via action diffusion},
  author={Chi, Cheng and Xu, Zhenjia and Feng, Siyuan and Cousineau, Eric and Du, Yilun and Burchfiel, Benjamin and Tedrake, Russ and Song, Shuran},
  journal={The International Journal of Robotics Research},
  pages={02783649241273668},
  year={2023},
  publisher={SAGE Publications Sage UK: London, England}
}

@misc{xue2024full,
    title={Full-Order Sampling-Based MPC for Torque-Level Locomotion Control via Diffusion-Style Annealing}, 
    author={Haoru Xue and Chaoyi Pan and Zeji Yi and Guannan Qu and Guanya Shi},
    year={2024},
    eprint={2409.15610},
    archivePrefix={arXiv},
    primaryClass={cs.RO},
    url={https://arxiv.org/abs/2409.15610}, 
}

@inproceedings{zhu2023efficient,
  title={Efficient object manipulation planning with monte carlo tree search},
  author={Zhu, Huaijiang and Meduri, Avadesh and Righetti, Ludovic},
  booktitle={2023 IEEE/RSJ International Conference on Intelligent Robots and Systems (IROS)},
  pages={10628--10635},
  year={2023},
  organization={IEEE}
}

@incollection{liu2010sampling,
  title={Sampling-based contact-rich motion control},
  author={Liu, Libin and Yin, KangKang and Van de Panne, Michiel and Shao, Tianjia and Xu, Weiwei},
  booktitle={ACM SIGGRAPH 2010 papers},
  pages={1--10},
  year={2010}
}

@article{aydinoglu2024consensus,
  title={Consensus complementarity control for multi-contact mpc},
  author={Aydinoglu, Alp and Wei, Adam and Huang, Wei-Cheng and Posa, Michael},
  journal={IEEE Transactions on Robotics},
  year={2024},
  publisher={IEEE}
}

@inproceedings{tassa2012synthesis,
  title={Synthesis and stabilization of complex behaviors through online trajectory optimization},
  author={Tassa, Yuval and Erez, Tom and Todorov, Emanuel},
  booktitle={2012 IEEE/RSJ International Conference on Intelligent Robots and Systems},
  pages={4906--4913},
  year={2012},
  organization={IEEE}
}

@article{kim2023contact,
  title={Contact-implicit mpc: Controlling diverse quadruped motions without pre-planned contact modes or trajectories},
  author={Kim, Gijeong and Kang, Dongyun and Kim, Joon-Ha and Hong, Seungwoo and Park, Hae-Won},
  journal={arXiv preprint arXiv:2312.08961},
  year={2023}
}

@article{kurtz2023inverse,
  title={Inverse dynamics trajectory optimization for contact-implicit model predictive control},
  author={Kurtz, Vince and Castro, Alejandro and {\"O}nol, Aykut {\"O}zg{\"u}n and Lin, Hai},
  journal={arXiv preprint arXiv:2309.01813},
  year={2023}
}

@article{heemels2000linear,
  title={Linear complementarity systems},
  author={Heemels, WPMH and Schumacher, Johannes M and Weiland, S},
  journal={SIAM journal on applied mathematics},
  volume={60},
  number={4},
  pages={1234--1269},
  year={2000},
  publisher={SIAM}
}

@article{anitescu1997formulating,
  title={Formulating dynamic multi-rigid-body contact problems with friction as solvable linear complementarity problems},
  author={Anitescu, Mihai and Potra, Florian A},
  journal={Nonlinear Dynamics},
  volume={14},
  number={3},
  pages={231--247},
  year={1997},
  publisher={Springer}
}

@article{contact_johnson_2019,
  title={Contact-implicit trajectory optimization using orthogonal collocation},
  author={Patel, Amir and Shield, Stacey Leigh and Kazi, Saif and Johnson, Aaron M and Biegler, Lorenz T},
  journal={IEEE Robotics and Automation Letters},
  volume={4},
  number={2},
  pages={2242--2249},
  year={2019},
  publisher={IEEE}
}

@article{posa2014direct,
  title={A direct method for trajectory optimization of rigid bodies through contact},
  author={Posa, Michael and Cantu, Cecilia and Tedrake, Russ},
  journal={The International Journal of Robotics Research},
  volume={33},
  number={1},
  pages={69--81},
  year={2014},
  publisher={Sage Publications Sage UK: London, England}
}

\enlargethispage{-0.5in}

\end{document}